\pdfoutput=1
\def\year{2016}
\documentclass[letterpaper]{article}
\usepackage{aaai16}
\usepackage{times}
\usepackage{helvet}
\usepackage{courier}

\usepackage{amsmath}    % Differentes fonctionnalites mathematiques
\usepackage{amsfonts}   % Fontes AMS 
\usepackage{amsmath}    % need for subequations
\usepackage{verbatim}   % useful for program listings
\usepackage{color}      % use if color is used in text
\usepackage[usenames, dvipsnames]{xcolor}
\usepackage{subfig}
\usepackage{tikz}
\usetikzlibrary{fit,calc,positioning,decorations.pathreplacing,matrix}
\usepackage{url}

\usepackage{amsthm}
\theoremstyle{definition}
\newtheorem{de}{Definition}%[chapter]
\newtheorem{ex}{Example}%[chapter]
\newtheorem*{continuancex}{Example \continuanceref ~Continued}
\newenvironment{continuance}[1]
  {\newcommand\continuanceref{\ref{#1}}\continuancex}
  {\endcontinuancex}

\theoremstyle{remark}

\renewcommand{\phi}{\varphi}

\begin{document}
% The file aaai.sty is the style file for AAAI Press 
% proceedings, working notes, and technical reports.
%
\title{Using Enthymemes to Fill the Gap between Logical Argumentation and Revision of Abstract Argumentation Frameworks}
\author{Jean-Guy Mailly\\Institute of Information Systems\\TU Wien, Autria\\jmailly@dbai.tuwien.ac.at}

\maketitle
\begin{abstract}
\begin{quote}
  In this paper, we present a preliminary work on an approach to fill
  the gap between logic-based argumentation and the numerous
  approaches to tackle the dynamics of abstract argumentation
  frameworks. Our idea is that, even when arguments and attacks are
  defined by means of a logical belief base, there may be some
  uncertainty about how accurate is the content of an argument, and so
  the presence (or absence) of attacks concerning it. We use
  enthymemes to illustrate this notion of uncertainty of arguments and
  attacks. Indeed, as argued in the literature, real arguments are
  often enthymemes instead of completely specified deductive
  arguments. This means that some parts of the pair (support, claim)
  may be missing because they are supposed to belong to some ``common
  knowledge'', and then should be deduced by the agent which receives
  the enthymeme.  But the perception that agents have of the common
  knowledge may be wrong, and then a first agent may state an
  enthymeme that her opponent is not able to decode in an accurate
  way. It is likely that the decoding of the enthymeme by the agent
  leads to mistaken attacks between this new argument and the existing
  ones. In this case, the agent can receive some information about
  attacks or arguments acceptance statuses which disagree with her
  argumentation framework. We exemplify a way to incorporate this new
  piece of information by means of existing works on the dynamics
  of abstract argumentation frameworks.
\end{quote}
\end{abstract}

\section{Introduction}
Argumentation frameworks (AFs) are a convenient way to represent conflicting information and to deduce which subset of the information can be inferred. For instance, they can be used to model dialogs between several agents \cite{AmgoudH06} or to analyze on-line discussion between social network users \cite{LeiteM11}. Argumentation can also be useful in a mono-agent setting, for instance to infer non-trivial conclusions from an inconsistent knowledge base \cite{BesnardHunter2001}.

The domain called {\em dynamics of argumentation} has become a hot topic in recent years, with numerous publications about it. The first ones consider really classical debate scenarios as the source of the dynamic process \cite{Boe09b,Boe09a,Cay10,BaumannB10,Bis11,Bisquert2013a,Bau12,BKTT13sum}. These approaches are perfectly well-suited for classical exchange of arguments between agents. Then, some approaches have proposed to consider new scenarios, closer to what happens with belief change in logical settings \cite{AGM85,KM91,KMUpdate}: these approaches propose to question the existing relation between arguments, and to modify this relation if it is required \cite{Doutre2014,NouiouaW14,Coste2014a,Coste2014b,Coste2015,BaumannB15,DillerIJCAI15}.\\

These works directly deal with the structure of the abstract AFs. An interesting question is ``What does AF revision mean when we consider logic-based AFs?''. Indeed, it is not obvious that attacks between arguments can be changed, since they stem from the logical inference relation; for instance, if arguments $a$ and $b$ attack each other because their claims are the negation of each other ({\em rebuttal} attack), then it is not accurate to consider that the attack between $a$ and $b$ could be removed. But this is only the case when we consider completely specified deductive arguments \cite{BesnardHunter2001}. As argued in the literature \cite{Hunter07}, the arguments which are used in real situations are often enthymemes, which are partially specified arguments: some parts of the support or some parts of the claim are not described, because it is supposed that they belong to some ``common knowledge''. There may be different reasons for an agent not to share some part of her knowledge, such as some cost on the communication process. Then, the agent who receives an enthymeme must decide how to complete the content of the enthymeme to be able to use it. But if the missing formulae to complete the enthymeme are not part of the agent's beliefs (or at least, are not considered by the agent as the most accurate way to complete the enthymeme), then she will use a badly completed enthymeme in her argumentation framework. We see with this situation that, even with an underlying logical belief base, the nature of arguments and attacks is not absolute; it depends on the agent's beliefs and on her way to complete enthymemes.\\

So we propose to consider the use of enthymemes in the argumentation process to explain the questionability of some attacks. We illustrate the possibility that a logic-based argumentation framework contains mistaken attacks. Then we show that the existing work on the dynamics of AFs can be used on such enthymeme-based AFs, as soon as a distinction between classical deductive arguments and enthymemes is done in the abstract AF, and that this distinction is used in the revision process.\\

The paper is organized as follows. The first section presents the background notions required to the understanding of the paper. In particular, we describe briefly belief revision, abstract argumentation and revision of AFs, and logic-based argumentation. Then in the second section, we focus on enthymemes in logic-based AFs; we explain how using enthymemes can be a source of mistaken attacks in the resulting AF. The following section illustrates the revision process on logic-based AFs which contain enthymemes. After the description of a basic approach in which each attack concerning an enthymeme is questionable, we propose a refinement of this approach based on the notion of fixed part of an enthymeme. Finally, the last section concludes the paper and sketches some interesting future work.

\section{Background}
\subsection{Belief Revision}
Belief revision is well-known when an agent's beliefs are represented in a logical setting. The intuitive idea is ``How can an agent incorporate a new piece of information into her beliefs?'', which is not a trivial question when the agent's previous beliefs and the new piece of information are conflicting. One of the most influencial works on this topic is the AGM framework \cite{AGM85}, which gives rationality postulates for belief change operators, when the beliefs are represented as deductively closed sets of formulae. Here we are interested in the adaptation of AGM revision to finite propositional logic by Katsuno and Mendelzon \shortcite{KM91}. They explain that revising a formula $\phi$ by a formula $\alpha$ is equivalent to selecting some models of $\alpha$ which are minimal w.r.t. some plausibility relation. This relation has to satisfy some properties.

\begin{de}[\citeauthor{KM91} \citeyear{KM91}]
A faithful assignment is a mapping from a formula $\phi$ to a total pre-order between interpretations $\leq_\phi$ such that:
\begin{enumerate}
  \item if $I \models \phi$ and $I' \models \phi$, then $I \simeq_\phi I'$;
  \item if $I \models \phi$ and $I' \not\models \phi$, then $I <_\phi I'$;
  \item if $\phi \equiv \phi'$, then $\leq_\phi = \leq_{\phi'}$.
\end{enumerate}
Then, a KM revision operator $\circ$ is a mapping from two formulae $\phi, \alpha$ to a new formula such that
\[
\mod(\phi \circ \alpha) = \min(\!\!\!\!\!\mod(\alpha),\leq_\phi)
\]
\end{de}

For instance, the Dalal revision operator can be defined through the pre-order built on the Hamming distance.

\begin{de}[\citeauthor{Hamming50} \citeyear{Hamming50}; \citeauthor{Dalal88} \citeyear{Dalal88}]
The {\em Hamming distance} between two propositional interpretations $I,I'$ is the number of assignments which differ between $I$ and $I'$, formally: $d_H(I,I') = |(I \backslash I') \cup (I' \backslash I)|$.\\
The total pre-order $\leq_\phi^{d_H}$ is defined by
\[
I \leq_\phi^{d_H} I' \text{ iff } \min_{J \in \!\!\!\mod\!(\phi)}(d_H(I,J)) \leq \min_{J \in \!\!\!\mod\!(\phi)}(d_H(I',J))
\]
The Dalal revision operator $\circ_{D}$ is a mapping from two formulae $\phi, \alpha$ to a new formula such that
\[
\mod(\phi \circ_{D} \alpha) = \min(\!\!\!\!\!\mod(\alpha),\leq_\phi^{d_H})
\]
\end{de}

Let us illustrate the behavior of the Dalal revision operator.

\begin{ex}
Consider $V = \{a,b,c,d\}$ and $\phi = [(a \wedge b) \vee (\neg a \wedge c) \vee \neg (b \vee (a \wedge c))] \wedge \neg d$. The models of $\phi$ are $\{\{a\},\{c\},\{a,b\},\{b,c\},\{a,b,c\}\}$. We revise $\phi$ by $\alpha = a \wedge \neg b \wedge c$. The models of $\alpha$ are $\{\{a,c\},\{a,c,d\}\}$. Table~\ref{tab:distanceModels} gives the Hamming distance between models of $\phi$ and models of $\alpha$.
\begin{table}[h]
\centering
\begin{tabular}{|c|c|c|}
  \hline
  & $\{a,c\}$ & $\{a,c,d\}$ \\ \hline
  $\emptyset$ & $2$ & $3$ \\ \hline
  $\{a\}$ & $1$ & $2$ \\ \hline
  $\{c\}$ & $1$ & $2$ \\ \hline
  $\{a,b\}$ & $2$ & $3$ \\ \hline
  $\{b,c\}$ & $2$ & $3$ \\ \hline
  $\{a,b,c\}$ & $1$ & $2$ \\ \hline
\end{tabular}
\caption{Hamming distance between models of $\phi$ and $\alpha$\label{tab:distanceModels}}
\end{table}

 Since the minimal Hamming distance between $\{a,c\}$ and a model of $\phi$ is $1$ ($d_H(\{a,c\},\{a\})$ for instance), while the distance between $\{a,c,d\}$ and any model of $\phi$ is at least $2$, then $\{a,c\} <_\phi^{d_H} \{a,c,d\}$, and so $\mod(\phi \circ_{Da} \alpha) = \{\{a,c\}\}$.
\end{ex}

\subsection{Abstract Argumentation and AF Revision}
An abstract AF is a directed graph which represents the arguments and the attacks between them. The usual problem to solve with such an abstract AF is ``How to determine which arguments are accepted?''. This question is tackled in the seminal paper by Dung \shortcite{Dung95}.

\begin{de}[\citeauthor{Dung95} \citeyear{Dung95}]
An argumentation framework (AF) is a pair $F = \langle A, R \rangle$ where $A$ is a set of abstract entities called {\em arguments}, and $R \subseteq A \times A$ is the {\em attack relation} which represents the conflicts between arguments.\\
Given a {\em semantics} $\sigma$, the $\sigma$-extensions of $F$, denoted $\sigma(F)$, are subsets of $A$ which can be accepted. An argument is then {\em skeptically accepted} by $F$ w.r.t. $\sigma$ iff it belongs to each $\sigma$-extension of $F$.\\

In this paper, we illustrate our approach on the {\em stable semantics}: $S \subseteq A$ is a {\em stable extension} of $F$ (denoted by $S \in st(F)$) iff
\begin{itemize}
  \item $\not\exists x, y \in S$ s.t. $(x,y) \in R$;
  \item $\forall y \in A \backslash S$, $\exists x \in S$ s.t. $(x,y) \in R$.
\end{itemize}
\end{de}

\begin{ex}
Given the set of arguments $A = \{x,y, z,\linebreak t,u\}$, the AF $F_1 = \langle A, R\rangle$ with $R = \{(x,y),(x,t),\linebreak (y,x),(y,z),(z,u),(t,u)\}$ is given in Figure~\ref{fig:ex-graphe}.

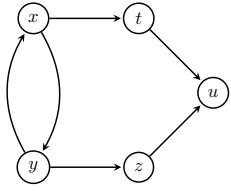
\begin{figure}[h!]
\centering
\scalebox{0.7}{
\begin{tikzpicture}[->,>=stealth,shorten >=1pt,auto,node distance=2cm,
                thick,main node/.style={circle,draw,font=\bfseries},scale=0.8]
\node[main node] (u) {$u$};
\node[main node] (z) [below left of=u] {$z$};
\node[main node] (t) [above left of=u] {$t$};
\node[main node] (x) [left of=t] {$x$};
\node[main node] (y) [left of=z] {$y$};

\path[->] (x) edge[bend left] (y) 
          (y) edge[bend left] (x)
          (x) edge (t)
          (y) edge (z)
          (t) edge (u)
          (z) edge (u);
\end{tikzpicture}}
\caption{The AF $F_1$ \label{fig:ex-graphe}}
\end{figure}

Its stable extensions are $st(F_1) = \{\{x,z\},\{y,t\}\}$.
\end{ex}

As explained in the introduction, the question of change in AFs has been tackled by several approaches. Here we use the {\em translation-based revision} from \cite{Coste2014b}. The idea of this method is to translate the AF and the semantics into a propositional formula, to use a KM revision operator to perform the expected change, and then to decode the models of the revised formula to obtain a set of revised AFs. The propositional encoding is a generalization of a result from Besnard and Doutre \shortcite{BesnardDoutre2004}. They have defined a formula $\Xi$, built on propositional variables corresponding to the arguments, such that the set of models of $\Xi$ exactly correspond to the set of stable extensions of an AF. \citeauthor{Coste2014b} \shortcite{Coste2014b} generalize this encoding with the addition of two other kinds of variables $V = \{att_{x,y} \mid x,y \in A\} \cup \{acc_x \mid x \in A\}$. $att_{x,y}$ means that there is an attack from the argument $x$ to the argument $y$, and $acc_x$ means that the argument $x$ is skeptically accepted.

\begin{de}[\citeauthor{Coste2014b} \citeyear{Coste2014b}]
Given an AF $F = \langle A = \{x_1,\dots,x_n\}, R \rangle$, the stable encoding of $F$ is
\[
f_{st}(F) = (\bigwedge_{(x,y)\in R} att_{x,y} )\wedge (\bigwedge_{(x,y)\notin R} \neg att_{x,y}) \wedge th_{st}(A)
\]
where
\[
\begin{array}{ll}
th_{st}(A) = & \bigwedge_{x \in A}[acc_x \Leftrightarrow \forall x_1, \dots, \forall x_n,\\
& (\bigwedge_{y \in A}(y \Leftrightarrow \bigwedge_{z \in A}(att_{z,y} \Rightarrow \neg z)) \Rightarrow x)]
\end{array}
\]
\end{de}

In general, $f_\sigma(F)$ can be defined for any semantics $\sigma$ as soon as the formula $\Xi$ exists; for semantics with a complexity higher than {\sf NP}, we can consider for instance QBF encodings to define $\Xi$.

Then, the revision operator is defined as follow:

\begin{de}[\citeauthor{Coste2014b} \citeyear{Coste2014b}]
Given $\circ$ a KM revision operator and $\phi$ a propositional formula built from the set of variables $V$, the {\em translation-based revision operator} $\star_\circ$ is defined as
\[
F \star_\circ \phi = dec(f_\sigma(F) \circ (\phi \wedge th_\sigma(A)))
\]
with $dec$ a mapping from a formula $\psi$ to a set of AFs $\mathcal{F}$ such that each AF $F' \in \mathcal{F}$ corresponds to one of the models $\omega$ of $\psi$: $(x,y)$ appears in $F'$ iff $att_{x,y}$ is true in $\omega$.
\end{de}

This general definition allows to change any attack and argument status as long as it is compatible with $\sigma$. If additional constraints should be satisfied,\footnote{Such as external constraint depending on the particular application, or some rules of the world.} the use of a constrained version is possible:

\begin{de}[\citeauthor{Coste2014b} \citeyear{Coste2014b}]
Given $\circ$ a KM revision operator and $\phi, \mu$ two propositional formulae built from the set of variables $V$, the {\em constrained translation-based revision operator} $\star_\circ^\mu$ is defined as
\[
F \star_\circ^\mu \phi = dec(f_\sigma(F) \circ (\phi \wedge th_\sigma(A) \wedge \mu))
\]
\end{de}

To conclude, let us mention a particular revision operator proposed by \cite{Coste2014b}: we call $\star_{att}$ (resp. $\star_{att}^\mu$) the translation-based (resp. constrained translation-based) revision operator which gives priority to the minimal change of the attack relation. This operator is similar to the Dalal-based revision revision operator $\star_{\circ_D}$, but it uses a weighted version of the Hamming distance such that changing the value of a single $att_{x,y}$ variable is more expensive than changing the value of each $acc_x$ variable.

\begin{ex}
We consider the AF $F_1$ given in Figure~\ref{fig:ex-graphe}. We suppose the existence of an integrity constraint $att_{t,u} \wedge att_{z,u}$, which means that the attacks from $t$ and $z$ to $u$ must not be removed. The result of the revision $F_1 \star_{\circ_D}^\mu acc_u$ is the AF $F_2$ described in Figure~\ref{fig:ex-constrained-revision}.
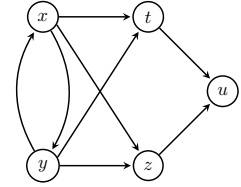
\begin{figure}[h!]
\centering
%\subfloat[$F_1$]{
\scalebox{0.7}{
\begin{tikzpicture}[->,>=stealth,shorten >=1pt,auto,node distance=2cm,
                thick,main node/.style={circle,draw,font=\bfseries},scale=0.8]
\node[main node] (u) {$u$};
\node[main node] (z) [below left of=u] {$z$};
\node[main node] (t) [above left of=u] {$t$};
\node[main node] (x) [left of=t] {$x$};
\node[main node] (y) [left of=z] {$y$};

\path[->] (x) edge[bend left] (y) 
          (y) edge[bend left] (x)
          (x) edge (t)
          (y) edge (z)
          (t) edge (u)
          (z) edge (u)
          (x.330) edge (z)
          (y.30) edge (t);
\end{tikzpicture}}
%}
\caption{$F_2 = F \star_{\circ_D}^\mu acc_u$ \label{fig:ex-constrained-revision}}
\end{figure}
Now the extensions are $st(F_2) = \{\{x,u\},\{y,u\}\}$, so $u$ is skeptically accepted.
\end{ex}

We focus on this kind of revision operators because \cite{Coste2014b}
already proposes a way to incorporate a constraint on the attack
relation, which is required by our approach. Other revision or update
operators could be used instead, but we should adapt their definition
to take into account the constraint.

\subsection{Logic-based Arguments: Deductive Arguments}
The question of the exact nature of arguments and attacks is tackled by several approaches which can be gathered under the name {\em structural argumentation}. Here we focus on one of the most prominent ones: deductive argumentation \cite{BesnardHunter2001}.
\begin{de}[\citeauthor{BesnardHunter2001} \citeyear{BesnardHunter2001}] \label{def:deductive-arg}
A deductive argument built from a belief base $\Delta$ is a pair $\langle \Phi, \alpha \rangle$, where $\Phi$ is called the {\em support} and $\alpha$ the {\em claim}, such that:
\begin{enumerate}
  \item $\Phi \subseteq \Delta$,
  \item $\Phi \not\vdash \bot$,
  \item $\Phi \vdash \alpha$,
  \item $\Phi$ is minimal with respect to $\subseteq$ among the sets of formulae which satisfy items 1. 2. and 3.
\end{enumerate}
\end{de}
There is an intuitive explanation to this definition. First the agent is supposed to use her beliefs to justify her claim, which explains the first condition. The second and third conditions guarantee that the claim is actually supported by the beliefs of the agent, but not by conflicting beliefs (for instance, the sentence ``It is raining and it is not raining, so I am the Queen of England.'' is not an argument at all). Finally, the last condition ensures that there is no useless piece of information in the support: ``It is raining, when it is raining I should use an umbrella, and I love chocolate. So I will use my umbrella.'' is not accurate either.\\

The conflicts between deductive arguments may have different natures. The most general sort of conflict is defined as follow:

\begin{de}[\citeauthor{BesnardHunter2001} \citeyear{BesnardHunter2001}]
A {\em defeater} for an argument $\langle \Phi, \alpha \rangle$ is an argument $\langle \Phi', \alpha'\rangle$ such that $\alpha' \vdash \neg(\phi_1 \wedge \dots \wedge \phi_n)$, for some $\{\phi_1,\dots,\phi_n\} \subseteq \Phi$.
\end{de}

It is possible to use deductive arguments to build an {\em argument tree} with the arguments and counterarguments which attack and defend a given claim.\\
We can also build a full argumentation framework from the set of deductive arguments generated from a belief base.

\begin{de}[\citeauthor{BesnardHunter2014} \citeyear{BesnardHunter2014}]
Given $A$ the set of deductive arguments generated from the belief base $\Delta$, the {\em exhaustive graph} associated with $\Delta$ is the AF\linebreak $F = \langle A, R \rangle$ with $R = \{(x,y) \in A \times A \mid x$ is a defeater for $y\}$.
\end{de}

Here we focus on the defeater relation, which is the most general one, but exhaustive graphs can be generated with another attack relation which guarantees additional properties for the defeaters (undercut, rebuttal, and so on). Moreover, these graphs may be infinite in general; Besnard and Hunter propose an approach to circumvent this problem. See \cite{BesnardHunter2014} for more details.

\section{Enthymemes and their Role in Mistaken Attacks}
\subsection{Intuitive Explanation}
Before formalizing our approach, we want to explain intuitively, with natural language arguments, why agents can disagree on the attack relation, and more generally why attacks could be questionable. Let us consider the following arguments:
\begin{description}
\item[(c)] The US army is preparing a secret plan to retreat from Afghanistan (source: Wikileaks).
\item[(b)] Our informed sources say that the Wikileaks documents are fake (source: NY Times).
\item[(a)] The media cannot be trusted on military issues (source: N. Chomsky).
\end{description}

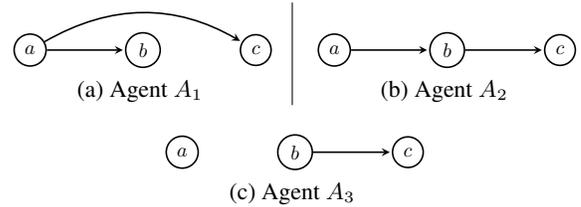
\begin{figure}[h!]
\begin{center}
\subfloat[Agent $A_1$\label{fig:agent-a1}]{
\scalebox{0.75}{
\begin{tikzpicture}[->,>=stealth,shorten >=1pt,auto,node distance=2cm,
                thick,main node/.style={circle,draw,font=\bfseries},scale=0.8]
\node[main node] (a) {$a$};
\node[main node] (b) [right of=a] {$b$};
\node[main node] (c) [right of=b] {$c$};

\path[->] (a) edge (b) 
    (a) edge[bend left] (c);
\end{tikzpicture}}
}
~
\vline
~
\subfloat[Agent $A_2$\label{fig:agent-a2}]{
\scalebox{0.75}{
\begin{tikzpicture}[->,>=stealth,shorten >=1pt,auto,node distance=2cm,
                thick,main node/.style={circle,draw,font=\bfseries},scale=0.8]
\node[main node] (a) {$a$};
\node[main node] (b) [right of=a] {$b$};
\node[main node] (c) [right of=b] {$c$};

\path[->] (a) edge (b) 
    (b) edge (c);
\end{tikzpicture}}
}\\
\subfloat[Agent $A_3$\label{fig:agent-a3}]{
\scalebox{0.75}{
\begin{tikzpicture}[->,>=stealth,shorten >=1pt,auto,node distance=2cm,
                thick,main node/.style={circle,draw,font=\bfseries},scale=0.8]
\node[main node] (a) {$a$};
\node[main node] (b) [right of=a] {$b$};
\node[main node] (c) [right of=b] {$c$};

\path[->] (b) edge (c);
\end{tikzpicture}}
}
\caption{\label{fig:chomsky-example} Three Agents Disagreement}
\end{center}
\end{figure}

Now we consider three agents $A_1, A_2, A_3$; each of them may have some personal beliefs which are not shared with the other agents.\\
\begin{itemize}
  \item $A_1$ thinks that Chomsky is the most credible source, and considers that Wikileaks is a media more reliable than NY Times. So her AF is the one given in Fig.~\ref{fig:agent-a1}.
  \item $A_2$ thinks that Chomsky is a more credible source than NY Times, and NY Times is a more credible source than Wikileaks. She also believes that Wikileaks cannot be seen as a media. So her AF is the one given in Fig.~\ref{fig:agent-a2}.
  \item Finally, $A_3$ thinks that NY Times is the most credible source, and that Chomsky is not reliable on this topic. So her AF is the one given in Fig.~\ref{fig:agent-a3}.
\end{itemize}
These personal AFs may depend on many different parameters (additional information which is not available to each agent, preferences, context, previous experience of each agent, and so on).

Of course, under the assumption that the agents share all their knowledge and beliefs, the personal beliefs of the agents can be represented as additional arguments and we obtain a single AF representing the whole information about a topic. But we think that this assumption is too strong for at least three reasons. First, there may be technical issues with this information sharing; for instance, there may be some cost on communication between agents, or the global amount of information in the network may be too important to be stored in a centralized way. Then, for strategical reasons, agents may {\em choose} not to share their knowledge and beliefs. Also, if argument are mined from natural language (for instance, for an analysis of social networks debates), there are likely some implicit pieces of information used in the argumentation process. This explains why some attacks may be questionable.

For instance, if the agents $A_1$ and $A_3$ consider that agent $A_2$ is trustworthy, then they could have to change the attack relation in their own AFs if they receive from agent $A_2$ the information ``$c$ should be accepted''. On the opposite, if agents vote to determine the arguments statuses, there will be a majority of agents ($A_1$ and $A_3$) voting against $c$ (meaning that $c$ is rejected in their AFs), so agent $A_2$ should modify the attack relation to incorporate this piece of information in her AF.

\paragraph{Enthymemes with partial support}
Now let us formalize this notion of ``arguments with partial knowledge'', and their role in the existence of mistaken attacks. Hunter \shortcite{Hunter07} defines what he calls {\em approximate arguments}, which are pairs $\langle \Phi, \alpha \rangle$ which do {\em not} satisfy the four conditions of deductive arguments. He classifies them depending on which properties they satisfy, and then focuses on enthymemes. An {\em enthymeme} is a pair $\langle \Phi, \alpha \rangle$ such that $\Phi \not\vdash \alpha$, but there is a set $\Psi \subseteq \Delta$ such that $\langle \Phi \cup \Psi, \alpha \rangle$ is a deductive argument. Intuitively, $\Psi$ represents some ``common knowledge'' that the agent supposes to be known by her opponents. Then it is not useful for the agent to state the full deductive argument to be able to exchange information and to reach her goal (persuading her opponent, helping to take a decision, negotiating, and so on).

\begin{ex}
To illustrate this concept, we borrow a simple example of real life use of enthymemes from \cite{Hunter07}. Let us consider John and his wife Yoko, who is going outside without an umbrella. If John tells her ``You should take your umbrella, because the weather report predicts rain'', there is no formal reason to consider that $\langle \Phi,\alpha \rangle$ (with $\Phi = \{rain\_predicted\}$ and $\alpha = take\_umbrella$) is an argument. It is in fact an enthymeme, because John supposes that $\Psi = \{rain\_predicted \Rightarrow take\_umbrella\}$ is part of the knowledge he shares with Yoko.
\end{ex}

One of the questions tackled in \cite{Hunter07} is ``How does the agent knows that $\Psi$ is {\em actually} part of the common knowledge?''. Hunter supposes that each agent has a way to evaluate the plausibility that a given formula will be part of the knowledge shared between her and another agent.

\begin{de}[\citeauthor{Hunter07} \citeyear{Hunter07}]
For each agent $A_i$ whose beliefs are expressed in the propositional language $\mathcal{L}$,
\begin{itemize}
  \item $\Delta_i \subseteq \mathcal{L}$ denotes her own {\em personal base},
  \item and for each other agent $A_j$, $\mu_{i,j}$ is a mapping from the language $\mathcal{L}$ to $[0,1]$, such that $\mu_{i,j}(\alpha)$ represents the certainty that $\alpha$ is common to both agents $A_i$ and $A_j$.
\end{itemize} 
\end{de}

\paragraph{On enthymemes and mistaken attacks} This mapping $\mu_{i,j}$ is used by the agent to build her arguments and decide whether they should be fully specified deductive arguments, or whether enthymemes can be used. The idea is simply to keep only the formulae $\phi$ in the support such that the associated value $\mu_{i,j}(\phi)$ is less than a given threshold $\tau$; the other ones can be omited because they are supposed to be known by agent $A_j$.

This process may lead to some problems in the exchange of arguments. There are at least two sources of mistakes.
\begin{enumerate}
  \item The mapping $\mu_{i,j}$ describes the {\em perception} that agent $A_i$ has of her common knowledge with $A_j$. If this perception is wrong, then there could be some exchange of enthymemes that the agent $A_j$ cannot decode accurately.
  \item Even with a good evaluation of the common knowledge by $\mu_{i,j}$, the choice of a bad threshold could also lead to enthymemes that the other agent cannot decode.
\end{enumerate}
In both these situations, agent $A_j$ receives some ``argument'' $a = \langle \Phi, \alpha \rangle$ which is not fully specified, and then she has to complete the support with some $\Psi'$ from her own belief base, which could of course lead to the addition of some attacks from an existing argument $b$ to this new argument $a$, for instance if the claim of $b$ is the formula $\neg \psi$, for some $\psi \in \Psi'$. Even if for low-level treatments, it can be represented as $a' = \langle \Phi \cup \Psi', \alpha \rangle$ (for instance to determine if possibly new incoming arguments attack it), at a higher level it is still the argument $a$ which is used. Indeed, this $a'$ is not an argument that $A_j$ has built by herself, since some of the premises are not part of her belief base.\footnote{To do this, it is a logical belief revision/expansion/update which should be performed, and this would likely have some side effects on the whole belief base, not only on the formulae involed in argument $a$.} Then, agent $A_j$ can receive a new piece of information about the argument $a$ which is incompatible with the attack from $b$ to $a$ (the simplest example being ``$a$ and $b$ should be accepted together''). So she has to build a new internal state $a''$ from a subset $\Psi''$ from her belief base; for the same reason as previously, at a higher level it is still the argument $a$ originally built from agent $A_i$'s beliefs.\\

\paragraph{Enthymemes with partial claim}
We have seen that enthymemes are a way to communicate arguments with partial support. Black and Hunter \shortcite{HunterBlack2012} also give some examples of enthymemes with a partial claim. Borrowing their example, let us consider the sentence $\alpha=$ ``John has bought The Times''. The enthymeme $\langle \{\alpha\}, \top \rangle$ can be interpreted in at least two ways, which lead to different claims:
\begin{enumerate}
  \item $\langle \{\alpha, \alpha \Rightarrow \beta\}, \beta \rangle$ with $\beta=$ ``John has bought a copy of the newspaper The Times'';
  \item $\langle \{\alpha, \alpha \Rightarrow \gamma\}, \gamma \rangle$ with $\gamma=$ ``John has bought the company which publishes the newspaper The Times''.
\end{enumerate}
Similarly to what we have described for enthymemes with a partial support, if an agent receives an argument $a$ which is in fact an enthymeme with a partial claim, some mistakes in the attack relation can appear. For instance, she may consider that $a$ attacks an argument $b$ because some part of $b$'s support is conflicting with the completed claim (either $\beta$ or $\gamma$ in our example). If she later receives a piece of information which is not compatible with this attack, then she may have to consider a removal of this attack (for instance, because the chosen claim is not accurate).\\

So we can formally define the class of enthymemes (with partial claims and partial supports) as follows:

\begin{de}[\citeauthor{HunterBlack2012} \citeyear{HunterBlack2012}]
Given $d = \langle \Phi, \alpha \rangle$ a deductive argument, an approximate argument $\langle \Phi', \alpha' \rangle$ is an enthymeme for $d$ iff $\Phi' \subset \Phi$ and $\alpha \vdash \alpha'$. 
\end{de}

Stated otherwise, the pair $\langle \Phi, \alpha \rangle$ is an enthymeme for $\langle \Phi\cup\Psi, \alpha \wedge \beta\rangle$, with $\Phi$ the partial support and $\alpha$ the partial claim. In the rest of this paper, we call such a pair $\langle \Phi, \alpha \rangle$ a {\em non-completed enthymeme} and $\langle \Phi\cup\Psi, \alpha \wedge \beta\rangle$ a {\em completed enthymeme}. A completed enthymeme may not satisfy the conditions stated in Definition~\ref{def:deductive-arg}, since the set of formulae $\Phi$ comes from another agent's belief base. Moreover, contrary to a fully specified argument stemming from the agent's beliefs, a completed enthymeme can be questioned.

\section{Dynamics of AFs and Enthymemes}
\subsection{Building a Dung's AF from Enthymemes}
For several reasons, the use of an abstract AF by the agent is interesting, even when she uses an underlying belief base. For instance, the developement of efficient approaches to solve abstract argumentation problems permits to obtain the conclusion of the agent's AF with respect to several semantics and inference policies (see for instance the competition of argumentation solvers \cite{ICCMA2015}). But to avoid the loss of information about the nature of arguments and attacks, we propose to refine the definition of the AF.

\begin{de}
Given $D$ and $E$ which denote respectively the agent's deductive arguments and enthymemes, the agent's {\em enthymeme-based AF} is $F(D,E) = \langle A, R \rangle$ with
\begin{itemize}
  \item $A = D \cup E$;
  \item $R = R_D \cup R_E$;
  \item $R_D \subseteq D \times D$ the set of certain attacks (between deductive arguments);
  \item $R_E \subseteq (A \times A) \backslash (D \times D)$ the set of questionable attacks (concerning at least one enthymeme).
\end{itemize}
\end{de}
Computing the extensions of such an AF is identical to the process for classical Dung's AFs; differentiating both kinds of attacks is useful only for the dynamics scenarios such as revision.

In this setting, each attack can be added or removed as soon as it concerns at least one enthymeme. We will refine this later.

\begin{ex}\label{ex:enthymeme-based-af}
Let $F_3$ be the enthymeme-based AF presented in Fig.~\ref{fig:enthymeme-based-af}. Arguments with rounded corners are the enthymemes while the other ones are deductive arguments. Similarly, the dashed arrows represent the questionable attacks, while the other ones are the certain attacks. In this example, we suppose that the agent has received the enthymemes in the following way:
\begin{itemize}
  \item $e_1 = \langle\{\alpha\},\gamma \rangle$, which has been completed by $\Psi_1 = \{\alpha\Rightarrow\beta,\beta\Rightarrow\gamma\}$;
  \item $e_2 = \langle \{\eta\},\top\rangle$, which has been completed by $\Psi_2 = \{\eta\Rightarrow\neg\epsilon\}$ in the support and $\neg\epsilon$ in the claim. 
\end{itemize}
\begin{figure}[h]
\centering
\scalebox{0.8}{
\begin{tikzpicture}[->,>=stealth,shorten >=1pt,auto,node distance=1.8cm,
                thick,deductive/.style={rectangle,draw,font=\bfseries},enthymeme/.style={rectangle,rounded corners=3pt,draw,font=\bfseries},scale=0.8]
\node[enthymeme] (a) {$e_1 = \langle\{\alpha,\alpha\Rightarrow\beta,\beta\Rightarrow\gamma\},\gamma \rangle$};
\node[deductive] (b) [below of=a] {$d_1 = \langle\{\delta,\delta\Rightarrow(\beta\wedge\neg\gamma)\}, \beta\wedge\neg\gamma\rangle$};
\node[deductive] (c) [below of=b] {$d_2 = \langle \{\epsilon,\epsilon\Rightarrow\neg\delta\},\neg\delta\rangle$};
\node[enthymeme] (d) [below of=c] {$e_2 = \langle \{\eta,\eta\Rightarrow\neg\epsilon\},\neg\epsilon\rangle$};

\path[->] (c) edge (b);
\path[->,dashed] (b) edge (a)
     (d) edge (c);
\end{tikzpicture}}
\caption{The Enthymeme-based AF $F_3$\label{fig:enthymeme-based-af}}
\end{figure}
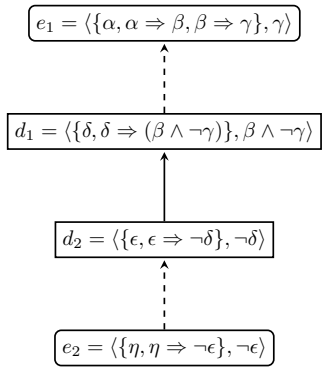
With this AF, the accepted arguments are $\{e_2,d_1\}$.
\end{ex}

\subsection{Applying Dynamics of Abstract AFs to Enthymeme-based AFs}
The existence of mistaken attacks in an enthymeme-based AF can be tackled through some approaches of the {\em dynamics of abstract argumentation} \cite{Bisquert2013a,Doutre2014,Coste2014a,Coste2014b,Coste2015}. In the case when some arguments and the relations between them are certain (in particular, when they are fully specified arguments instead of enthymemes), integrity constraints can simply be added to these revision/update/enforcement operators to ensure that forbidden attacks will not be added, and mandatory attacks will not be removed. Since it is already defined by \cite{Coste2014b}, we will exemplify the dynamics of argumentation with their {\em constrained revision} approach, presented previously. We can encode an integrity constraint to fix the attacks and non-attacks concerning the deductive arguments into the setting from \cite{Coste2014b}.

\begin{de}
Given $F(D,E)$ an enthymeme-based AF, the integrity constraint on deductive arguments is
\[
\mu_D = (\bigwedge_{(x,y) \in R_D} att_{x,y}) \wedge (\bigwedge_{(x,y) \in (D\times D)\backslash R_D} \neg att_{x,y})
\]
\end{de}

Now, if the agent receives some piece of information about the arguments statuses or the attack relation, then she can use the AF revision operator $\star_{att}^{\mu_D}$ as defined previously in the case when this new piece of information disagrees with the current AF. This revision operator guarantees that the relations between deductive arguments will not be modified during the revision process, which is desirable since they are directly stemming from the logical inference relation. 

\begin{continuance}{ex:enthymeme-based-af}
We continue the previous example. The agent receives the piece of information ``$e_1$ should be accepted'', which corresponds to the formula $acc_{e_1}$. The integrity constraint is $att_{d_2,d_1} \wedge \neg att_{d_1,d_2}$, which ensures that the attacks between the deductive arguments $d_1$ and $d_2$ will not be modified. The possible results are given in Fig.~\ref{fig:possible-results-revision}.
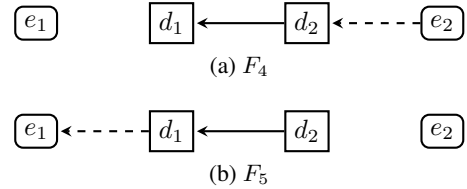
\begin{figure}[h!]
\begin{center}
\subfloat[$F_4$\label{fig:result1}]{
\begin{tikzpicture}[->,>=stealth,shorten >=1pt,auto,node distance=1.8cm,
                thick,deductive/.style={rectangle,draw,font=\bfseries},enthymeme/.style={rectangle,rounded corners=3pt,draw,font=\bfseries},scale=0.8]
\node[enthymeme] (a) {$e_1$};
\node[deductive] (b) [right of=a] {$d_1$};
\node[deductive] (c) [right of=b] {$d_2$};
\node[enthymeme] (d) [right of=c] {$e_2$};

\path[->] (c) edge (b);
\path[->,dashed] (d) edge (c);
\end{tikzpicture}
}\\
\subfloat[$F_5$\label{fig:result2}]{
\begin{tikzpicture}[->,>=stealth,shorten >=1pt,auto,node distance=1.8cm,
                thick,deductive/.style={rectangle,draw,font=\bfseries},enthymeme/.style={rectangle,rounded corners=3pt,draw,font=\bfseries},scale=0.8]
\node[enthymeme] (a) {$e_1$};
\node[deductive] (b) [right of=a] {$d_1$};
\node[deductive] (c) [right of=b] {$d_2$};
\node[enthymeme] (d) [right of=c] {$e_2$};

\path[->] (c) edge (b);
\path[->,dashed] (b) edge (a);
\end{tikzpicture}
}
\caption{\label{fig:possible-results-revision} Possible Results of the Revision}
\end{center}
\end{figure}
\end{continuance}

The exact {\em change operator} which should be used depends on the the properties expected for the process, for instance it is well-known that performing an update \cite{Bisquert2013a,Doutre2014} is accurate when the change is explained by an evolution of the world, while performing a revision \cite{Coste2014a,Coste2014b} is accurate when the evolution only concerns the agent's beliefs about the world; thus these operations do not satisfy the same properties. Similarly, among the different approaches in the state of the art about the dynamics of AFs, each of them do not have the same expressivity. For instance, the revision approach described in \cite{Coste2014a} permits to revise by a formula concerning the extensions, while the translation-based approach illustrated here permits to revise by a formula concerning skeptical acceptance of arguments and attacks at the same time.
So, the choice of a change operator completely depends on the application and the agent's needs and preferences. In the following, we continue to consider revision to be consistent with the previous example, but update and extension enforcement \cite{BaumannB10} could be considered as well.

\subsection{From Revised AFs to new Completed Enthymemes}
After obtaining the result of the revision process, the agent should now decode this result to determine which of the revised AFs is the most plausible real AF corresponding to her beliefs, and which enthymemes should be internally modified (and {\em how} they should be internally modified) to ensure that the abstract AF and the logic-based AF coincide.

\begin{de}
Let $\mathcal{F}$ be the set of AFs obtained from the revision
process. For each $F' \in \mathcal{F}$, $F'$ is called an {\em
  acceptable AF} iff for each attack which differs between the
original AF $F$ and $F'$, the agent's belief base contains some
formulae which allow to complete the enthymemes s.t. this new
completion is consistent with the attacks in $F'$.
\end{de}

\begin{continuance}{ex:enthymeme-based-af}
Continuing the previous example, let us suppose that the agent's belief base contains the formulae $\Psi'=\{\alpha\Rightarrow\theta,\theta\Rightarrow\gamma\}$. Then the enthymeme $e_1$ can be completed into $\langle\{\alpha,\alpha\Rightarrow\theta,\theta\Rightarrow\gamma\},\gamma \rangle$, which leads to the acceptable AF $F_4$ given in Fig.~\ref{fig:acceptable1}. Similarly, if the agent's belief base contains the formulae $\Psi'' = \{\eta\Rightarrow\iota\}$, then the agent can consider the acceptable AF $F_5$ given in Fig.~\ref{fig:acceptable2}, since $e_2$ can be completed into $\langle \{\eta,\eta\Rightarrow\iota\},\iota\rangle$.
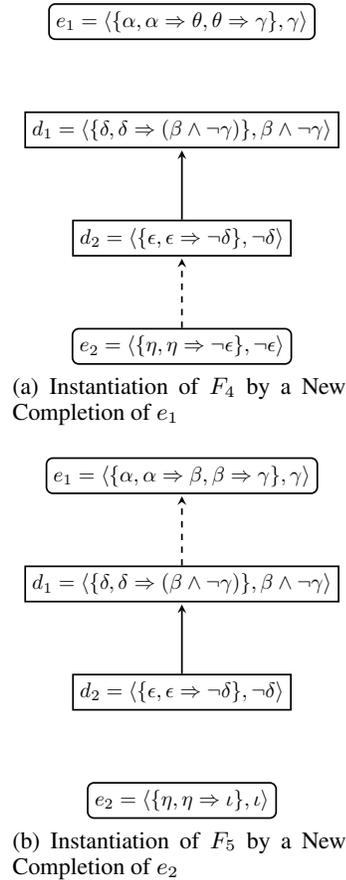
\begin{figure}[h]
\centering
\subfloat[Instantiation of $F_4$ by a New Completion of $e_1$\label{fig:acceptable1}]{
\scalebox{0.8}{
\begin{tikzpicture}[->,>=stealth,shorten >=1pt,auto,node distance=1.8cm,
                thick,deductive/.style={rectangle,draw,font=\bfseries},enthymeme/.style={rectangle,rounded corners=3pt,draw,font=\bfseries},scale=0.8]
\node[enthymeme] (a) {$e_1 = \langle\{\alpha,\alpha\Rightarrow\theta,\theta\Rightarrow\gamma\},\gamma \rangle$};
\node[deductive] (b) [below of=a] {$d_1 = \langle\{\delta,\delta\Rightarrow(\beta\wedge\neg\gamma)\}, \beta\wedge\neg\gamma\rangle$};
\node[deductive] (c) [below of=b] {$d_2 = \langle \{\epsilon,\epsilon\Rightarrow\neg\delta\},\neg\delta\rangle$};
\node[enthymeme] (d) [below of=c] {$e_2 = \langle \{\eta,\eta\Rightarrow\neg\epsilon\},\neg\epsilon\rangle$};

\path[->] (c) edge (b);
\path[->,dashed] (d) edge (c);
\end{tikzpicture}}
}

\subfloat[Instantiation of $F_5$ by a New Completion of $e_2$\label{fig:acceptable2}]{
\scalebox{0.8}{
\begin{tikzpicture}[->,>=stealth,shorten >=1pt,auto,node distance=1.8cm,
                thick,deductive/.style={rectangle,draw,font=\bfseries},enthymeme/.style={rectangle,rounded corners=3pt,draw,font=\bfseries},scale=0.8]
\node[enthymeme] (a) {$e_1 = \langle\{\alpha,\alpha\Rightarrow\beta,\beta\Rightarrow\gamma\},\gamma \rangle$};
\node[deductive] (b) [below of=a] {$d_1 = \langle\{\delta,\delta\Rightarrow(\beta\wedge\neg\gamma)\}, \beta\wedge\neg\gamma\rangle$};
\node[deductive] (c) [below of=b] {$d_2 = \langle \{\epsilon,\epsilon\Rightarrow\neg\delta\},\neg\delta\rangle$};
\node[enthymeme] (d) [below of=c] {$e_2 = \langle \{\eta,\eta\Rightarrow\iota\},\iota\rangle$};

\path[->] (c) edge (b);
\path[->,dashed] (b) edge (a);
\end{tikzpicture}}
}
\caption{Different Acceptable AFs\label{fig:acceptable-revised-afs}}
\end{figure}
\end{continuance}

When the set of acceptable AFs is not a singleton, we can consider two different solutions:
\begin{itemize}
  \item the agent can keep the whole set as the result, to express the uncertainy of the result of the revision, as suggested by \cite{Bisquert2013a,Coste2014a,Coste2014b,Doutre2014} which consider that revising or updating an AF can lead to a set of AFs;
  \item the agent can use external information (preferences between AFs, preferences between formulae in the enthymemes, and so on) to select a single acceptable AF as the result.
\end{itemize}
None of them is in general more desirable than the other one, the choice depends on the situation (specific application, user's preferences, computational issues,$\dots$).
% Of course, the choice of one of these approaches depends on the application, none of them is in general more desirable than the other one.

\subsection{Refining Questionable Attacks}
In the previous parts, we suppose that each attack concerning an enthymeme is questionable. But we can be more precise in the definition of the enthymeme-based AF. Indeed, even when we consider an enthymeme $e$, some of the attacks concerning it may be certain. We know that some parts of $e$, that we have called previously the partial support and the partial claim, are fixed. If the reason of an attack between $e$ and a deductive argument is a logical conflict involving one of these fixed parts of $e$, then this attack can be considered as certain. Similarly, when we consider another enthymeme $e'$, if there is an attack between $e$ and $e'$ which is stemming from the fixed part of $e$ and the fixed part of $e'$, then this attack cannot be removed either.\\

Let us first formalize this notion of fixed part.

\begin{de}
If $a = \langle \Phi,\alpha\rangle$ is a deductive argument or a non-completed enthymeme, then the {\em fixed part} of $a$ is $fix(a) = \Phi \cup \{\alpha\}$.\\
If $a = \langle \Phi \cup \Psi,\alpha\wedge\beta\rangle$ is a completed enthymeme, then $fix(a) = \Phi \cup \{\alpha\}$.
\end{de}

So if we consider a fully specified deductive argument or a non-completed enthymeme, the fixed part is the set of all the formulae involved in it. But when we consider an enthymeme {\em completed with the agent's beliefs}, then the fixed part is the set of formulae which appear in the enthymeme that the agent has originally received, but do not appear in the completed version of it.

\begin{continuance}{ex:enthymeme-based-af}
Let us consider again the arguments $d_1$, $d_2$, $e_1$ and $e_2$. The fixed parts of the deductive arguments are trivially the union of their support and their claim.\\
The result is more interesting for the enthymemes:
\begin{itemize}
  \item $fix(e_1) = \{\alpha,\gamma\}$;
  \item $fix(e_2) = \{\eta,\top\}$;
\end{itemize}
\end{continuance}

Now let us define the involved part of an argument in an attack.

\begin{de}
Let $a = \langle \Phi,\alpha\rangle$ and $b = \langle \Phi',\alpha'\rangle$ be two arguments (deductive arguments, completed enthymemes or non-completed enthymemes). If there is an attack between $a$ and $b$, then the {\em involved part} of $a$ in the conflict between $a$ and $b$, denoted by $inv_b(a)$, is the set $\Psi \subseteq \Phi \cup \{\alpha\}$ such that $(\bigwedge_{\psi \in \Psi} \psi) \wedge (\bigwedge_{\phi' \in \Phi' \cup \{\alpha'\}}\phi') \vdash \bot$ and $\Psi$ is minimal w.r.t. $\subseteq$. Otherwise, $inv_b(a) = inv_a(b) = \emptyset$.
\end{de}

So, if we have a rebuttal conflict between $a$ and $b$ (meaning that the claims are the contradiction of each other) then $inv_b(a) = \{\alpha\}$ and $inv_a(b) = \{\alpha'\}$. If the conflict is an undercut from $a$ to $b$ (meaning that the claim $\alpha$ of $a$ is conflicting with some part $\phi'$ of the support of $b$), then $inv_b(a) = \{\alpha\}$ and $inv_a(b) = \{\phi'\}$. 

\begin{continuance}{ex:enthymeme-based-af}
Now we can see which parts of the arguments $d_1$, $d_2$, $e_1$ and $e_2$ are involved in conflicts. The certain attack $(d_2,d_1)$ comes from the contradiction between $\delta$ and $\neg \delta$, so $inv_{d_1}(d_2) = \{\neg\delta\}$ and $inv_{d_2}(d_1) = \{\delta\}$. Concerning the questionable attacks, we have:
\begin{itemize}
  \item $inv_{e_1}(d_1) = \{\beta\wedge\neg\gamma\}$ and $inv_{d_1}(e_1) = \{\beta\Rightarrow\gamma\}$;
  \item $inv_{e_2}(d_2) = \{\epsilon\}$ and $inv_{d_2}(e_2) = \{\neg\epsilon\}$.
\end{itemize}
\end{continuance}

Now we can refine the definition of an enthymeme-based AF.

\begin{de}
Given $D$ and $E$ which denote respectively the agent's deductive arguments and enthymemes, the agent's {\em refined enthymeme-based AF} is $F(D,E) = \langle A, R \rangle$ with
\begin{itemize}
  \item $A = D \cup E$;
  \item $R = R_C \cup R_Q$;
  \item $R_C = \{(x,y) \in A \times A \mid inv_y(x) \subseteq fix(x) \text{ and } inv_x(y) \subseteq fix(y)\}$: the set of certain attacks;
  \item $R_Q \subseteq (A \times A) \backslash R_C$: the set of questionable attacks.
\end{itemize}
We use $R_D$ as a notation for $R_C \cap (D \times D)$, which is the set of attacks between deductive arguments.
\end{de}

Of course, if an argument is a fully specified deductive argument, then the part of it which is involved in conflicts is a fixed part. So to refine the AF, we need to check if it is the case with the enthymemes.

\begin{continuance}{ex:enthymeme-based-af}
Studying the relations between involved parts and fixed parts for the enthymemes $e_1$ and $e_2$, we obtain the following:
\begin{itemize}
  \item $inv_{d_1}(e_1) = \{\beta\Rightarrow\gamma\} \not\subseteq fix(e_1) = \{\alpha,\gamma\}$;
  \item $inv_{d_2}(e_2) = \{\neg\epsilon\} \not\subseteq fix(e_2) = \{\eta,\top\}$.
\end{itemize}
So none of the attacks $(e_2,d_2)$ and $(d_1,e_1)$ is certain.
\end{continuance}

But we can exhibit more interesting cases, for which the use of a refined enthymeme-based AF leads to another result than the basic enthymeme-based AF.

\begin{ex}
Let $d_3 = \langle \{\nu,\nu\Rightarrow\neg\lambda\},\neg\lambda\rangle$ be a deductive argument, and $e_3 = \langle \{\kappa\},\lambda\rangle$ an enthymeme, which can be completed for instance by the additional support $\Phi' = \{\kappa\Rightarrow\lambda\}$.\\
It is easy to see here that $inv_{d_3}(e_3) = \{\lambda\} \subseteq fix(e_3) = \{\kappa,\lambda\}$, so the conflict between $d_3$ and $e_3$ is not questionable, and the AF corresponding to these arguments is $F_6$ given in Fig.~\ref{fig:refined-AF}.
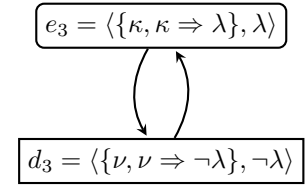
\begin{figure}[h]
\centering
\begin{tikzpicture}[->,>=stealth,shorten >=1pt,auto,node distance=1.8cm,
                thick,deductive/.style={rectangle,draw,font=\bfseries},enthymeme/.style={rectangle,rounded corners=3pt,draw,font=\bfseries},scale=0.8]
\node[enthymeme] (a) {$e_3 = \langle\{\kappa,\kappa\Rightarrow\lambda\},\lambda \rangle$};
\node[deductive] (b) [below of=a] {$d_3 = \langle\{\nu,\nu\Rightarrow\neg\lambda\}, \neg\lambda\rangle$};

\path[->] (a) edge[bend right] (b)
          (b) edge[bend right] (a);
\end{tikzpicture}
\caption{The Refined Enthymeme-based AF $F_6$\label{fig:refined-AF}}
\end{figure}
\end{ex}

When we consider these refined enthymeme-based AFs in the revision process, the integrity constraint must be adapted to take into account each certain attack, and not only the ones between deductive arguments:

\begin{de}
Given $F(D,E)$ a refined enthymeme-based AF, the integrity constraint on certain attacks is
\[
\mu_C = (\bigwedge_{(x,y) \in R_C} att_{x,y}) \wedge (\bigwedge_{(x,y) \in (D \times D)\backslash R_D} \neg att_{x,y})
\]
\end{de}

This new constraint ensures that a certain attack will not be removed during the revision process, and that attacks between deductive arguments will not be added if they do not belong to the original AF.

\subsection{Back to Chomsky Example}
To conclude, let us formalize the intuitive ``Chomsky example'', showing the different completions of enthymemes which lead to the different agents AFs. We use the following propositional variables: $retreat$ means that the US army will retreat; $wkr$ means that the {\bf W}i{\bf k}ileaks information about {\bf r}etreat is true; $wkf$ means that the {\bf W}i{\bf k}ileaks documents are {\bf f}ake; $mnt$ means that {\bf m}edia can {\bf n}ot be {\bf t}rusted on military issues. As they are stated, the arguments $a$, $b$ and $c$ which are shared by the agents are these ones:
\[
a = \langle\{mnt\},\top\rangle, b = \langle\{wkf\},\top\rangle, c = \langle\{wkr\},\top\rangle
\]
% \begin{itemize}
%   \item $a = \langle\{mnt\},\top\rangle$;
%   \item $b = \langle\{wkf\},\top\rangle$;
%   \item $c = \langle\{wkr\},\top\rangle$.
% \end{itemize}
All of them are enthymemes. For all the agents, the completion of $c$ is $\langle\{wkr,wkr \Rightarrow retreat\},retreat\rangle$. But they disagree on the completion of the other enthymemes.
Agent $A_1$ considers that $a = \langle \{mnt, mnt \Rightarrow \neg wkf, mnt \Rightarrow \neg wkr\}, \neg wkf \wedge \neg wkr\rangle$ and $b = \langle \{wkf\},\top\rangle$.
Agent $A_2$ completes the enthymemes as follows: $a = \langle \{mnt, mnt \Rightarrow \neg wkf\}, \neg wkf\rangle$ and $b = \langle \{wkf, wkf \Rightarrow \neg wkr\},\neg wkr\rangle$.

Finally, agent $A_3$ uses these completions of enthymemes: $a = \langle \{mnt\}, \top\rangle$ and $b = \langle \{wkf, wkf \Rightarrow \neg wkr\},\neg wkr\rangle$.

These completions of enthymemes lead to the AFs described in
Figure~\ref{fig:chomsky-example}, with all arguments which are
enthymemes, and all attacks which are questionable. So here, in case
of a revision, the revision operator is used with the integrity
constraint $\top$, which is equivalent to a revision without a
constraint.\\

\begin{figure}[h!]
\centering
\scalebox{0.75}{
\begin{tikzpicture}[->,>=stealth,shorten >=1pt,auto,node distance=1.8cm,
                thick,deductive/.style={rectangle,draw,font=\bfseries},enthymeme/.style={rectangle,rounded corners=3pt,draw,font=\bfseries},scale=0.8]
\node[enthymeme] (a) {$a = \langle\{mnt, mnt \Rightarrow \neg wkf\}, \neg wkf\rangle$};

\node[enthymeme] (b) at (5,0) {$b$};
\node[enthymeme] (c) [right of=b] {$c$};

\path[->,dashed] (a) edge (b);
\end{tikzpicture}}
%}
\caption{Revision for Agent $A_1$\label{fig:revision-chomsky-a1}}
\end{figure}
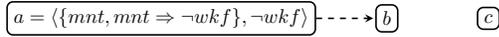

We mentioned in the introduction two scenarios which require to use dynamics of argumentation techniques. First, we suppose that agent $A_2$ is considered to be trustworthy by other agents. Then, when she says that $c$ should be accepted (which is represented by the formula $acc_c$), the other agents have to revise their AF with this new piece of information. The result of the revision for $A_1$, with a corresponding completion of enthymemes which are modified because of the revision, is given in Figure~\ref{fig:revision-chomsky-a1}.

\begin{figure}[h!]
\centering
\subfloat{
\scalebox{0.7}{
\begin{tikzpicture}[->,>=stealth,shorten >=1pt,auto,node distance=3cm,
                thick,deductive/.style={rectangle,draw,font=\bfseries},enthymeme/.style={rectangle,rounded corners=3pt,draw,font=\bfseries},scale=0.8]
\node[enthymeme] (a) {$a$};
\node[enthymeme] (b) [right of=a] {$b = \langle \{wkf\},\top\rangle$};
\node[enthymeme] (c) [right of=b] {$c$};

\end{tikzpicture}}
}

\rule{4cm}{0.2mm}

\subfloat{
\scalebox{0.8}{
\begin{tikzpicture}[->,>=stealth,shorten >=1pt,auto,node distance=1.8cm,
                thick,deductive/.style={rectangle,draw,font=\bfseries},enthymeme/.style={rectangle,rounded corners=3pt,draw,font=\bfseries},scale=0.8]
\node[enthymeme] (a) {$a = \langle\{mnt, mnt \Rightarrow \neg wkf\}, \neg wkf\rangle$};
\node[enthymeme] (b) at (5,0) {$b$};
\node[enthymeme] (c) [right of=b] {$c$};

\path[->,dashed] (a) edge (b) 
    (b) edge (c);
\end{tikzpicture}}
}
\caption{Possible Revisions for Agent $A_3$\label{fig:revision-chomsky-a3}}
\end{figure}
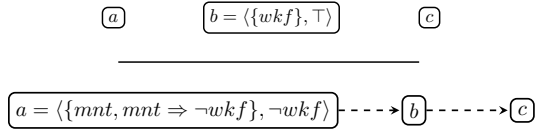

Similarly, for $A_3$, Figure~\ref{fig:revision-chomsky-a3} describes the possible revised AFs, with the modified enthymemes corresponding to it.

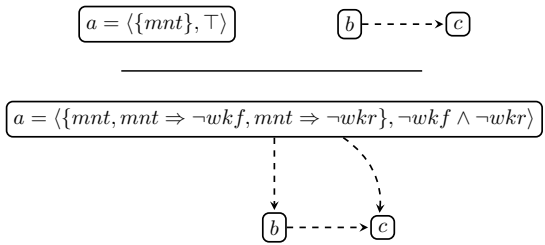
\begin{figure}[h!]
\centering
\subfloat{
\scalebox{0.8}{
\begin{tikzpicture}[->,>=stealth,shorten >=1pt,auto,node distance=1.8cm,
                thick,deductive/.style={rectangle,draw,font=\bfseries},enthymeme/.style={rectangle,rounded corners=3pt,draw,font=\bfseries},scale=0.8]
\node[enthymeme] (a) {$a = \langle \{mnt\}, \top \rangle$};
\node[enthymeme] (b) at (4,0) {$b$};
\node[enthymeme] (c) [right of=b] {$c$};

\path[->,dashed] (b) edge (c);
\end{tikzpicture}}
}

\rule{4cm}{0.2mm}

\subfloat{
\scalebox{0.8}{
\begin{tikzpicture}[->,>=stealth,shorten >=1pt,auto,node distance=1.8cm,
                thick,deductive/.style={rectangle,draw,font=\bfseries},enthymeme/.style={rectangle,rounded corners=3pt,draw,font=\bfseries},scale=0.8]
\node[enthymeme] (a) {$a = \langle\{mnt, mnt \Rightarrow \neg wkf, mnt \Rightarrow \neg wkr\}, \neg wkf \wedge \neg wkr\rangle$};
\node[enthymeme] (b) [below of=a] {$b$};
\node[enthymeme] (c) [right of=b] {$c$};

\path[->,dashed] (a) edge (b) 
    (b) edge (c)
    (a) edge[bend left] (c);
\end{tikzpicture}}
}
\caption{Possible Revisions for Agent $A_2$\label{fig:revision-chomsky-a2}}
\end{figure}

Finally, the other scenario was a vote on the acceptance status of
$c$. Since the majority of agents rejects $c$, $A_2$ has to revise her
AF by $\neg acc_c$ to find an agreement with the majority. Possible
results are described in Figure~\ref{fig:revision-chomsky-a2}.

\section{Conclusion}
In this paper, we argue that, in realistic situations, agents do not
share {\em all} their knowledge and beliefs. There are different
possible reasons, among them, technical and strategical reasons seem
to be the most intuitive explanations. Also, implicit information is
frequently used in natural language argumentation (on social networks
for instance). When this situation occurs, there is some uncertainty
in the resulting argumentation frameworks built by the agents. It is
likely that agents' opinion about arguments' meaning and relations
between arguments will differ; there may be some misunderstanding in
the communication process which leads to mistakes in the generation of
arguments and attacks. Here, this is formalized with the use of
enthymemes, instead of deductive arguments, to represent the uncertain
nature of some arguments. For this reason, the reception of a new
piece of information (supposed to be reliable) can force the agents to
question the current attack relation to obtain a result which is
compatible with the new piece of information.
We have described formally how the use of enthymemes in the
argumentation process can lead to the existence of these mistaken
attacks, and how to define an argumentation framework which makes the
distinction between the certain attacks and the questionable
attacks. Then, we have seen that the existing works on the dynamics of
abstract AFs can be used to perform a change on an enthymeme-based AF
when it is required to incorporate a new piece of information. Here,
we exemplify it with the translation-based revision from
\cite{Coste2014b}, but it can be adapted to any revision, update or
enforcement approach as soon as it is possible to consider an
integrity constraint on the attack relation.\\

This paper only presents some preliminary work on this question. Many
future works can be envisioned. First, we want to model the
uncertainty by other means than enthymemes. For instance, using
weights could lead to the definition of original change operators
which define the notion of minimal change w.r.t. these weights; it
would then be more expensive to change a single attack which has a
high weight than to change several attacks with low weights. Determining
what can be the origin of these weights is particularly
interesting. Combined with the use of enthymemes, we think that giving
a low weight to an attack which is easy to modify (because there are
many possible completions of enthymemes in the belief base) is an
interesting way to tackle the problem.
For the EAFs defined in this paper, as well as the weighted approach
mentioned above, several questions are opened. We have made
the simplifying hypothesis that the agents will have some possible
completion of arguments at their disposal, which is not always the
case in real world situations. Similarly, the revision operators may
lead to an empty-set of results (because of the integrity constraint),
or on the opposite, to a non-singleton set of results. All these cases
must be investigated to ensure the possibility of some practical
applications.
The complexity of revising such framework, compared to the original
revision approach in the abstract setting, will also be studied to be
able to identify which approaches can be used for real applications.
We also plan to use some of the existing pieces of
software for the dynamics of AFs (in particular the one described in
\cite{Coste2015,Wallner2015} for extension enforcement), to study the
scalability of the approaches on practical examples. But it requires
first an important work to build argumentation graphs from logical
knowledge-bases, since the existing works focus only on argumentation
trees \cite{EfstathiouHunter2008,EfstathiouHunter2011,BesnardGPR2010}.

% Uncomment if accepted
\section{Acknowledgments}
This work as been supported by the Austrian Science Fund (FWF) under
grants P25521 and I1102.

%%%%%%%%%%%%%%%% Bibliography
% \bibliographystyle{aaai}
% \bibliography{revision_enthymemes}

\begin{thebibliography}{}

\bibitem[\protect\citeauthoryear{Alchourr{\'o}n, G{\"a}rdenfors, and
  Makinson}{1985}]{AGM85}
Alchourr{\'o}n, C.~E.; G{\"a}rdenfors, P.; and Makinson, D.
\newblock 1985.
\newblock On the logic of theory change : Partial meet contraction and revision
  functions.
\newblock {\em Journal of Symbolic Logic} 50:510--530.

\bibitem[\protect\citeauthoryear{Amgoud and Hameurlain}{2006}]{AmgoudH06}
Amgoud, L., and Hameurlain, N.
\newblock 2006.
\newblock An argumentation-based approach for dialog move selection.
\newblock In {\em Proc. of {ArgMAS} 2006},  128--141.

\bibitem[\protect\citeauthoryear{Baumann and Brewka}{2010}]{BaumannB10}
Baumann, R., and Brewka, G.
\newblock 2010.
\newblock Expanding argumentation frameworks: Enforcing and monotonicity
  results.
\newblock In {\em Proc. of COMMA 2010},  75--86.

\bibitem[\protect\citeauthoryear{Baumann and Brewka}{2015}]{BaumannB15}
Baumann, R., and Brewka, G.
\newblock 2015.
\newblock {AGM} meets abstract argumentation: Expansion and revision for {D}ung
  frameworks.
\newblock In {\em Proc. of IJCAI 2015}.

\bibitem[\protect\citeauthoryear{Baumann}{2012}]{Bau12}
Baumann, R.
\newblock 2012.
\newblock What does it take to enforce an argument? minimal change in abstract
  argumentation.
\newblock In {\em Proc. of ECAI 2012},  127--132.

\bibitem[\protect\citeauthoryear{Besnard and Doutre}{2004}]{BesnardDoutre2004}
Besnard, P., and Doutre, S.
\newblock 2004.
\newblock Checking the acceptability of a set of arguments.
\newblock In {\em Proc. of NMR 2004},  59--64.

\bibitem[\protect\citeauthoryear{Besnard and Hunter}{2001}]{BesnardHunter2001}
Besnard, P., and Hunter, A.
\newblock 2001.
\newblock A logic-based theory of deductive arguments.
\newblock {\em Artificial Intelligence} 128(1-2):203--235.

\bibitem[\protect\citeauthoryear{Besnard and Hunter}{2014}]{BesnardHunter2014}
Besnard, P., and Hunter, A.
\newblock 2014.
\newblock Constructing argument graphs with deductive arguments: A tutorial.
\newblock {\em Argument and Computation} 5(1):5--30.

\bibitem[\protect\citeauthoryear{Besnard \bgroup et al\mbox.\egroup
  }{2010}]{BesnardGPR2010}
Besnard, P.; Gr\'egoire, E.; Piette, C.; and Raddaoui, B.
\newblock 2010.
\newblock {MUS}-based generation of arguments and counter-arguments.
\newblock In {\em Proc. of IRI 2010},  239--244.

\bibitem[\protect\citeauthoryear{Bisquert \bgroup et al\mbox.\egroup
  }{2011}]{Bis11}
Bisquert, P.; Cayrol, C.; de~Saint-Cyr, F.~D.; and Lagasquie-Schiex, M.-C.
\newblock 2011.
\newblock Change in argumentation systems: Exploring the interest of removing
  an argument.
\newblock In {\em Proc. of SUM 2011},  275--288.

\bibitem[\protect\citeauthoryear{Bisquert \bgroup et al\mbox.\egroup
  }{2013}]{Bisquert2013a}
Bisquert, P.; Cayrol, C.; de~Saint{-}Cyr, F.~D.; and Lagasquie{-}Schiex, M.
\newblock 2013.
\newblock Enforcement in argumentation is a kind of update.
\newblock In {\em Proc. of {SUM} 2013},  30--43.

\bibitem[\protect\citeauthoryear{Black and Hunter}{2012}]{HunterBlack2012}
Black, E., and Hunter, A.
\newblock 2012.
\newblock A relevance-theoretic framework for constructing and deconstructing
  enthymemes.
\newblock {\em Journal of Logic and Computation} 22(1):55--78.

\bibitem[\protect\citeauthoryear{Boella, Kaci, and van~der
  Torre}{2009a}]{Boe09b}
Boella, G.; Kaci, S.; and van~der Torre, L.
\newblock 2009a.
\newblock Dynamics in argumentation with single extensions: Abstraction
  principles and the grounded extension.
\newblock In {\em Proc. of ECSQARU 2009},  107--118.

\bibitem[\protect\citeauthoryear{Boella, Kaci, and van~der
  Torre}{2009b}]{Boe09a}
Boella, G.; Kaci, S.; and van~der Torre, L.
\newblock 2009b.
\newblock Dynamics in argumentation with single extensions: Attack refinement
  and the grounded extension.
\newblock In {\em Proc. of AAMAS 2009},  1213--1214.

\bibitem[\protect\citeauthoryear{Booth \bgroup et al\mbox.\egroup
  }{2013}]{BKTT13sum}
Booth, R.; Kaci, S.; Rienstra, T.; and van~der Torre, L.
\newblock 2013.
\newblock A logical theory about dynamics in abstract argumentation.
\newblock In {\em Proc. of SUM 2013}. Springer.
\newblock  148--161.

\bibitem[\protect\citeauthoryear{Cayrol, de Saint-Cyr, and
  Lagasquie-Schiex}{2010}]{Cay10}
Cayrol, C.; de~Saint-Cyr, F.~D.; and Lagasquie-Schiex, M.-C.
\newblock 2010.
\newblock Change in abstract argumentation frameworks: Adding an argument.
\newblock {\em Journal of Artificial Intelligence Research} 38:49--84.

\bibitem[\protect\citeauthoryear{Coste-Marquis \bgroup et al\mbox.\egroup
  }{2014a}]{Coste2014a}
Coste-Marquis, S.; Konieczny, S.; Mailly, J.-G.; and Marquis, P.
\newblock 2014a.
\newblock On the revision of argumentation systems: Minimal change of arguments
  statuses.
\newblock In {\em Proc. of KR 2014},  72--81.

\bibitem[\protect\citeauthoryear{Coste-Marquis \bgroup et al\mbox.\egroup
  }{2014b}]{Coste2014b}
Coste-Marquis, S.; Konieczny, S.; Mailly, J.-G.; and Marquis, P.
\newblock 2014b.
\newblock A translation-based approach for revision of argumentation
  frameworks.
\newblock In {\em Proc. of JELIA 2014},  77--85.

\bibitem[\protect\citeauthoryear{Coste-Marquis \bgroup et al\mbox.\egroup
  }{2015}]{Coste2015}
Coste-Marquis, S.; Konieczny, S.; Mailly, J.-G.; and Marquis, P.
\newblock 2015.
\newblock Extension enforcement in abstract argumentation as an optimization
  problem.
\newblock In {\em Proc. of IJCAI 2015},  2876--2882.

\bibitem[\protect\citeauthoryear{Dalal}{1988}]{Dalal88}
Dalal, M.
\newblock 1988.
\newblock Investigations into a theory of knowledge base revision: Preliminary
  report.
\newblock In {\em Proc. of AAAI 1988},  475--479.

\bibitem[\protect\citeauthoryear{Diller \bgroup et al\mbox.\egroup
  }{2015}]{DillerIJCAI15}
Diller, M.; Haret, A.; Linsbichler, T.; R{\"u}mmele, S.; and Woltran, S.
\newblock 2015.
\newblock An extension-based approach to belief revision in abstract
  argumentation.
\newblock In {\em Proc. of IJCAI 2015},  2926--2932.

\bibitem[\protect\citeauthoryear{Doutre, Herzig, and
  Perrussel}{2014}]{Doutre2014}
Doutre, S.; Herzig, A.; and Perrussel, L.
\newblock 2014.
\newblock A dynamic logic framework for abstract argumentation.
\newblock In {\em Proc. of {KR} 2014},  62--71.

\bibitem[\protect\citeauthoryear{Dung}{1995}]{Dung95}
Dung, P.~M.
\newblock 1995.
\newblock On the acceptability of arguments and its fundamental role in
  nonmonotonic reasoning, logic programming, and n-person games.
\newblock {\em Artificial Intelligence} 77(2):321--357.

\bibitem[\protect\citeauthoryear{Efstathiou and
  Hunter}{2008}]{EfstathiouHunter2008}
Efstathiou, V., and Hunter, A.
\newblock 2008.
\newblock Algorithms for effective argumentation in classical propositional
  logic : A connection graph approach.
\newblock In {\em Proc. of FoIKS 2008},  272--290.

\bibitem[\protect\citeauthoryear{Efstathiou and
  Hunter}{2011}]{EfstathiouHunter2011}
Efstathiou, V., and Hunter, A.
\newblock 2011.
\newblock Algorithms for generating arguments and counterarguments in
  propositional logic.
\newblock {\em International Journal of Approximate Reasoning} 52(6):675--704.

\bibitem[\protect\citeauthoryear{Hamming}{1950}]{Hamming50}
Hamming, R.~W.
\newblock 1950.
\newblock Error detecting and error correcting codes.
\newblock {\em Bell System Technical Journal} 29(2):147--160.

\bibitem[\protect\citeauthoryear{Hunter}{2007}]{Hunter07}
Hunter, A.
\newblock 2007.
\newblock Real arguments are approximate arguments.
\newblock In {\em Proc. of {AAAI}'07},  66--71.

\bibitem[\protect\citeauthoryear{Katsuno and Mendelzon}{1991}]{KM91}
Katsuno, H., and Mendelzon, A.~O.
\newblock 1991.
\newblock Propositional knowledge base revision and minimal change.
\newblock {\em Artificial Intelligence} 52:263--294.

\bibitem[\protect\citeauthoryear{Katsuno and Mendelzon}{1992}]{KMUpdate}
Katsuno, H., and Mendelzon, A.~O.
\newblock 1992.
\newblock On the difference between updating a knowledge base and revising it.
\newblock In G{\"a}rdenfors, P., ed., {\em Belief Revision}.
\newblock  183--203.

\bibitem[\protect\citeauthoryear{Leite and Martins}{2011}]{LeiteM11}
Leite, J., and Martins, J.
\newblock 2011.
\newblock Social abstract argumentation.
\newblock In {\em Proc. of {IJCAI} 2011},  2287--2292.

\bibitem[\protect\citeauthoryear{Nouioua and W{\"{u}}rbel}{2014}]{NouiouaW14}
Nouioua, F., and W{\"{u}}rbel, E.
\newblock 2014.
\newblock Removed set-based revision of abstract argumentation frameworks.
\newblock In {\em Proc. of ICTAI'14},  784--791.

\bibitem[\protect\citeauthoryear{Thimm and Villata}{2015}]{ICCMA2015}
Thimm, M., and Villata, S.
\newblock 2015.
\newblock First {I}nternational {C}ompetition on {C}omputational {M}odels of
  {A}rgumentation ({ICCMA'15}).
\newblock see \url{http://argumentationcompetition.org/2015/}.

\bibitem[\protect\citeauthoryear{Wallner, Niskanen, and
  J\"arvisalo}{2015}]{Wallner2015}
Wallner, J.~P.; Niskanen, A.; and J\"arvisalo, M.
\newblock 2015.
\newblock Complexity results and algorithms for extension enforcement in
  abstract argumentation.
\newblock In {\em Proc. of AAAI'15}.

\end{thebibliography}
%\input{revision_enthymemes.bbl}

\end{document}